\documentclass[letterpaper, 10 pt, conference]{ieeeconf}  
\usepackage[utf8]{inputenc}
\usepackage{mathtools}
\usepackage{amsmath}
\usepackage{amssymb }
\usepackage{graphicx}
\usepackage{amsfonts}
\usepackage{algorithm}
\usepackage{algpseudocode}
\usepackage{multirow}
\usepackage{array}
\usepackage{pifont}
\usepackage{booktabs}
\usepackage{xcolor}     
\usepackage{subfig}
\usepackage{cite}

\DeclareMathAlphabet{\mathcal}{OMS}{cmsy}{m}{n}
\DeclareSymbolFont{largesymbols}{OMX}{cmex}{m}{n}

\IEEEoverridecommandlockouts                              

\usepackage{epsfig} 
\usepackage{mathptmx} 
\usepackage{times} 
\usepackage{amsmath} 
\usepackage{amssymb}  
\usepackage{graphicx} 
\usepackage{url}

\title{\LARGE \bf
AirSwarm: Enabling Cost-Effective Multi-UAV Research\\ with COTS drones
}

\author{Xiaowei~Li$^{\sim}$,
        Kuan~Xu$^{\sim}$,
        Fen Liu,
        Ruofei Bai,
        Shenghai~Yuan$^*$,~\IEEEmembership{Member,~IEEE}
        and~Lihua~Xie,~\IEEEmembership{Fellow,~IEEE}
\thanks{ $^{\sim}$ Equal Contribution.  $^{*}$ Corresponding Author.}
\thanks{This work is supported by the National Research Foundation of Singapore under its Medium-Sized Center for Advanced Robotics Technology Innovation.}
\thanks{All authors are with the Centre for Advanced Robotics Technology Innovation (CARTIN), School of Electrical and Electronic Engineering, Nanyang Technological University, 50 Nanyang Avenue, Singapore 639798, { \{shyuan,elhxie\}@ntu.edu.sg}.}
}

\begin{document}

\maketitle
\thispagestyle{empty}
\pagestyle{empty}

\begin{abstract}
Traditional unmanned aerial vehicle (UAV) swarm missions rely heavily on expensive custom-made drones with onboard perception or external positioning systems, limiting their widespread adoption in research and education. To address this issue, we propose AirSwarm. AirSwarm democratizes multi-drone coordination using low-cost commercially available drones such as Tello or Anafi, enabling affordable swarm aerial robotics research and education. Key innovations include a hierarchical control architecture for reliable multi-UAV coordination, an infrastructure-free visual SLAM system for precise localization without external motion capture, and a ROS-based software framework for simplified swarm development. Experiments demonstrate cm-level tracking accuracy, low-latency control, communication failure resistance, formation flight, and trajectory tracking. By reducing financial and technical barriers, AirSwarm makes multi-robot education and research more accessible. The complete instructions and open source code will be available at \url{https://github.com/vvEverett/tello_ros}.
\end{abstract}

\textbf{\textit{Index Terms} --- Drone Swarms, Multi-Robot Systems, SLAM, Low-Cost Robotics}


\section{Introduction}

Unmanned Aerial Vehicle (UAV) swarm systems have shown great potential in applications such as collaborative inspection \cite{lyu2023spins,cao2025cooperative,lyu2022structure,xu2024cost}, goods delivery \cite{zhu2024swarm,li2023autotrans,ji2022robust}, and field surveys \cite{xu2024d,yuan2014Autonomous}. They offer better scalability and resilience, ensuring redundancy and fault tolerance in dynamic environments. However, their adoption in research and education is severely limited by high hardware costs and system complexity \cite{zhou2022swarm}. Worse still, regulations often work against academic and research efforts, making swarm research on UAVs extremely difficult to advance.

Existing swarm research \cite{cao2025cooperative,li2023autotrans, 
yin2023decentralized,
xu2024d,
zhu2024swarm,
zhou2022swarm,
gao2022meeting} is heavily biased toward custom-built drones \cite{Zhang2024UniQuad,sa2017build,baca2021mrs,foehn2022agilicious,pan2023canfly} that rely on DIY hardware and firmware, often requiring hundreds of hours for development, integration, and procurement. Although these drones demonstrate impressive capabilities, their implementation is highly resource-intensive and demands expertise across multiple domains, including aerodynamics, embedded systems, computer vision, and networking.  The fragmented nature of development \cite{chen2024cost} not only slows progress but also creates a high barrier to entry for researchers and educators who lack specialized knowledge in all these areas. In contrast, commercial off-the-shelf
(COTS) drones like the DJI Mavic series offer limited API support, while options like the DJI Tello and Anafi lack robust perception, restricting their use in scalable swarm applications. A balance between accessibility, modularity, and computational capability is crucial for advancing UAV swarm research.

A key challenge in expanding UAV swarm research is enabling precise state estimation and real-time coordination on affordable COTS drones by effectively integrating sensor feedback with control systems \cite{esfahani2018new,wang2017heterogeneous,esfahani2020unsupervised}. Traditional methods rely on costly external localization, such as motion capture systems \cite{kushleyev2013towards,mohta2018fast} or RTKGPS/UWB-based solutions \cite{hauert2011reynolds}, which, while effective, significantly limit accessibility and scalability. High-precision platforms like the Flying Machine Arena \cite{lupashin2014platform} and Crazyswarm \cite{preiss2017crazyswarm} have demonstrated impressive swarm control, but their reliance on expensive infrastructure restricts their use to well-funded research institutions, preventing broader adoption in real-world applications.

\begin{figure}
    \centering
    \includegraphics[width=1\linewidth]{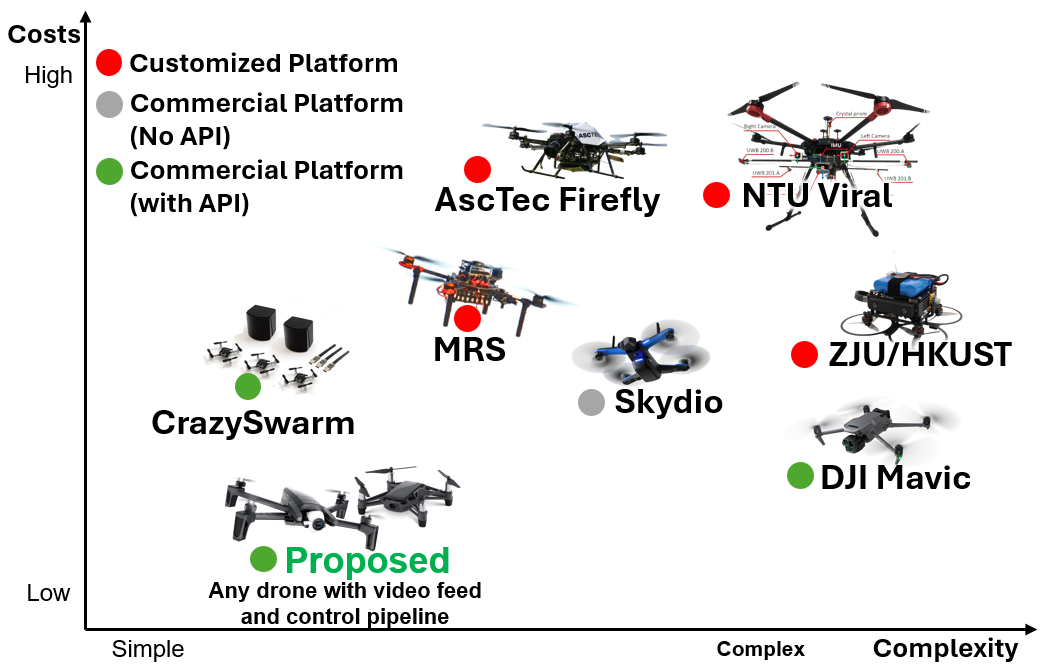}
    \caption{Comparison of Swarm systems by cost and complexity, highlighting the proposed approach.}
    \label{fig:Motivation}
    \vspace{-15pt}
\end{figure}

\begin{figure*}
    \centering
    \includegraphics[width=1\linewidth]{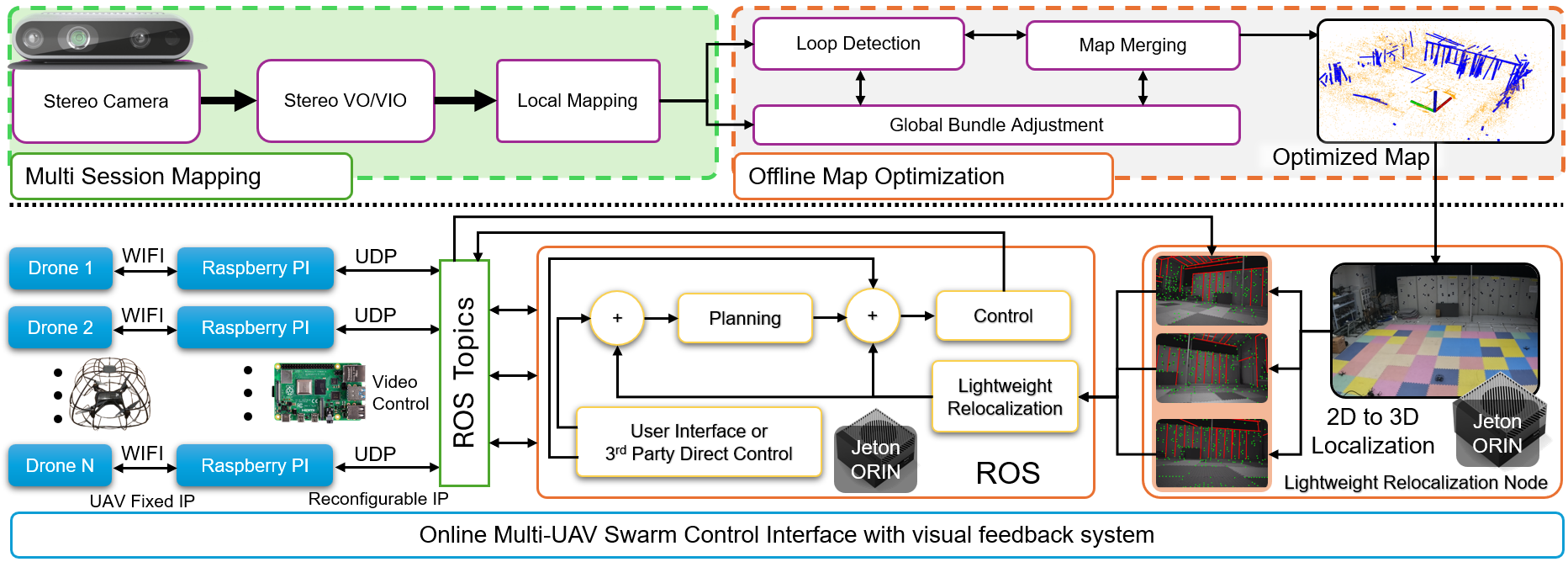}
    \caption{AirSwarm System Architecture. The diagram shows the complete workflow from environmental sensing to drone control, including: multi-session mapping, hardware communication architecture, and the integrated control interface.}
    \label{fig:system-overview}
    \vspace{-15pt}
\end{figure*}

To address external sensing issues, onboard alternatives, such as  LIO \cite{zhu2024swarm} or VIO \cite{xu2024d}, have a higher chance of enabling a COTS drone swarm with onboard autonomy. However, they come with their own challenges, including intermittent communication, drift accumulation, sensor calibration complexities, and susceptibility to environmental factors such as lighting conditions and electromagnetic interference. Overall, a swarm, in a nutshell, presents a complex set of challenges that need to be balanced.

To address these challenges, we present AirSwarm, a novel swarm architecture designed for low-cost, scalable UAV research using commercial off-the-shelf (COTS) drones, such as DJI Tello or Parrot Anafi. The system incorporates Raspberry Pi units to manage IP address conflicts and compress video streams for efficient data handling.
Each UAV localizes independently using visual SLAM-based prior mapping, without associating with past observations, ensuring robustness against intermittent network connectivity. A PD controller is implemented for precise UAV control, alongside a dedicated interface for streamlined swarm management.
We demonstrate that AirSwarm enhances the success rate of existing control and planning algorithms, making it a practical and accessible solution for swarm robotics research. Our contributions can be summarized as follows:

\begin{itemize}
    \item \textbf{Resilient Multi-UAV Control with Noisy and Conflicting Networks:} We propose a network-adaptive communication framework that mitigates IP-conflict of the COTS drones, ensuring robust swarm coordination in real-world wireless conditions with intermittent noises.

    \item \textbf{Low-Cost, Scalable Swarm Research Platform with COTS Drones:} Introduces a cost-effective, infrastructure-free swarm system utilizing commercial off-the-shelf (COTS) drones, such as DJI Tello or Parrot Anafi, combined with lightweight localization and hierarchical control architecture, lowering the barrier for swarm robotics research.

    \item \textbf{Open-Source, Reproducible Swarm Research Platform:} Provides a ROS-based, open-source framework with detailed deployment instructions, hardware integration guidelines, and real-world experimental validation, making swarm research more accessible, scalable, and reproducible for academia and industry \url{https://github.com/vvEverett/tello_ros}. 

\end{itemize}

The significance of this work lies in democratizing UAV swarm research by reducing cost and complexity barriers, enabling broader institutional participation in robotics research while advancing practical, infrastructure-independent swarm operations.

\vspace{-15pt}
\section{Related Works} \label{sec:related_works}
UAV swarm research has made significant progress, yet key challenges persist in achieving scalable, cost-effective, and flexible deployments \cite{cao2020online}. One major limitation is the reliance on expensive external localization systems \cite{xia2024av-dtec,lei2025audio,liang2025unsupervised,yang2024av,yuan2024mmaud}, restricting accessibility and real-world applicability. Traditional approaches such as motion capture \cite{kushleyev2013towards, mohta2018fast} and RTK-GPS/UWB solutions \cite{hauert2011reynolds} provide high-precision tracking but entail substantial financial and infrastructural costs. Advanced swarm control has been demonstrated in systems like FMA \cite{lupashin2014platform} and Crazyswarm \cite{preiss2017crazyswarm}, but their dependence on costly VICON motion capture and other external positioning systems restricts use to well-funded institutions. Alternative solutions like ICARUS \cite{lieser2016low} attempt to lower costs through optical tracking but remain confined to controlled indoor environments \cite{cao2022direct}. Overcoming these limitations is essential for expanding UAV swarm accessibility and real-world deployment.

Many UAV swarm studies focus on custom-built drones \cite{cao2025cooperative, li2023autotrans, yin2023decentralized, xu2024d, zhu2024swarm, zhou2022swarm, gao2022meeting}, which, while capable, require extensive development and specialized expertise. This fragmented approach \cite{chen2024cost} slows progress and creates high entry barriers for researchers lacking expertise in aerodynamics, embedded systems, and networking. In contrast, commercial off-the-shelf (COTS) drones offer a more accessible alternative but suffer from limited API support and weak onboard perception, constraining their scalability in swarm applications. A balance between accessibility, modularity, and computational capability is crucial for progress.

A promising solution for enabling swarm applications on COTS drones is robust onboard state estimation \cite{chen2024ig,esfahani2020local}, where Visual Simultaneous Localization and Mapping (SLAM) is vital \cite{esfahani2021learning}. Traditional SLAM methods such as ORB-SLAM3 \cite{campos2021orb} and PL-SLAM \cite{gomez2019pl} improved localization robustness, while learning-based methods like DROID-SLAM \cite{teed2021droid,chen2024salient,yang2022overcoming,ji2024lio,deng2024incremental,yin2024outram,cao2024mopa} offer resilience but demand high computational resources. Moreover, multi-UAV SLAM \cite{schmuck2019ccm} \cite{jiarui2023cpslam}   systems often struggle with calibration and intermittent data streaming, making deployment challenging.

Hybrid SLAM architectures \cite{xu2023airvo,cai2025bev,yuan2024large,Nguyen2025ULOC} have emerged as an optimal solution. AirVO \cite{xu2023airvo} combines learning-based feature detection with classical optimization, improving illumination robustness while maintaining efficiency. Its successor, AirSLAM \cite{xu2024airslam}, further enhances loop closure detection and map reuse through a unified point-line feature network, making it particularly suitable for UAVs with limited onboard processing power.

\section{AirSwarm System} \label{sec:swarmlite}

\subsection{System Overview}

The AirSwarm system employs a hierarchical architecture that integrates perception and control capabilities within a tiered framework, as illustrated in Fig~\ref{fig:system-overview}. The system comprises three primary functional layers:

\begin{itemize}
    \item \textbf{Mapping Subsystem:} Implements multi-session mapping using stereo cameras, integrating stereo visual-inertial odometry for initial pose estimation and local mapping with point-line features. The process includes loop detection, global bundle adjustment, and offline optimization to generate an optimized environmental model.
    
    \item \textbf{Communication Architecture:} Establishes a centralized network topology where COTS drones with onboard monocular cameras and IMUs communicate via WiFi to Raspberry Pi units. These units serve as bridges, utilizing a fixed-to-reconfigurable IP architecture that interfaces with ROS Topics through Ethernet connections to the central processor.
    
    \item \textbf{Control Framework:} It implements a versatile control stack compatible with various COTS drones equipped with video feedback for planning and control functions with lightweight relocalization. The relocalization pipeline processes monocular image streams for 2D-to-3D localization against the pre-built map, supporting both direct user interface control and programmatic access through multiple computing platforms. 
\end{itemize}

The NVIDIA Jetson AGX Orin functions as the computational nexus, executing mapping and localization algorithms while coordinating the drone network through a centralized architecture. This design enables infrastructure-independent operation via visual SLAM localization while maintaining efficient command distribution through hierarchical communication. This approach optimizes computational resource utilization while providing the unified coordination necessary for precise multi-UAV formation control.

\subsection{ROS-Based Universal COTS Framework}

Based on ROS and COTS SDK, we implemented  a drone control system with carefully designed architecture for simplicity and usability. Our key design considerations include: 
\begin{enumerate}
    \item A unified architecture that enables seamless transition between single and multiple drone operations;
    \item Thread-safe implementations for video stream processing to ensure reliable real-time performance;
    \item A comprehensive yet minimalist API that encapsulates complex flight control functionalities through simple ROS topics;
    \item A streamlined configuration approach where users only need to specify basic parameters like drone ID and IP address.
\end{enumerate}

These design choices significantly lower the technical barrier for robotics research and education, particularly in multi-drone applications. The system's modular architecture also facilitates straightforward integration with SLAM algorithms, making it a versatile platform for both research exploration and educational practices in swarm robotics.

\subsection{Drift-Free Visual Localization}

To achieve drift-free localization \cite{esfahani2019towards} for multiple UAVs using only low-cost cameras, we first build an accurate point-line map using a stereo camera based on AirSLAM \cite{xu2024airslam}, and then perform relocalization within this map using onboard monocular cameras .
Our relocalization consists of four steps. First, line and point features are extracted using a unified convolutional neural network (CNN). Then, a bag-of-words (BoW) vocabulary is utilized to retrieve keyframes within the map. Subsequently, feature matching between retrieved keyframes and the query frame is performed using a graph neural network (GNN). Finally, the Perspective-n-Point (PnP) algorithm is applied to estimate the camera pose.

To fully utilize the computational resources of the Jetson platform, we use both GPU and CPU to perform relocalization . Our feature detection and matching are executed on the GPU, while similar keyframe retrieval and pose estimation are performed on the CPU. This design enables our visual-only localization to achieve both the robustness and accuracy of learning-based methods while maintaining near real-time efficiency on embedded platforms.

\subsection{Unified Multi-UAV Architecture}

To facilitate research and educational applications, we developed a comprehensive control system that manages both single-drone operation and multi-drone fleet coordination.
Let $n \in \mathbb{N^+}$ denote the number of drones in the system, where $n=1$ represents single-drone operation and $n>1$ indicates fleet configuration. For each drone $i \in \{1,2,\ldots,n\}$.
Let $\mathbf{p}_i = [x_i, y_i, z_i, \psi_i]^T \in \mathbb{R}^4$ denote the 3D position and yaw angle, $\mathbf{v}_i = [v_{x,i}^b, v_{y,i}^b, v_{z,i}^b, \omega_{\psi,i}]^T \in \mathbb{R}^4$ represent the body-frame velocity command, and $\mathbf{p}^{d}_{i} = [x^{d}_{i}, y^{d}_{i}, z^{d}_{i}, \psi^{d}_{i}]^T \in \mathbb{R}^4$ indicate the desired state. The error vector 
$\mathbf{e}_i = [e_{x,i}, e_{y,i}, e_{z,i}, e_{\psi,i} ]^T \in \mathbb{R}^4$
contains position errors, yaw error in world-frame, while $\eta_i \in \mathbb{R}$ represents the battery status.
The system state matrices are:
\begin{align}
\mathbf{P} &= [\mathbf{p}_1, \mathbf{p}_2, \ldots, \mathbf{p}_n]^T \in \mathbb{R}^{n \times 4}, \\
\mathbf{V} &= [\mathbf{v}_1, \mathbf{v}_2, \ldots, \mathbf{v}_n]^T \in \mathbb{R}^{n \times 4}, \\
\mathbf{P}^{d} &= [\mathbf{p}^{d}_{1}, \mathbf{p}^{d}_{2}, \ldots, \mathbf{p}^{d}_{n}]^T \in \mathbb{R}^{n \times 4}, \\
\mathbf{E} &= [\mathbf{e}_1, \mathbf{e}_2, \ldots, \mathbf{e}_n]^T \in \mathbb{R}^{n \times 4}, \\
\boldsymbol{\eta} &= [\eta_1, \eta_2, \ldots, \eta_n]^T \in \mathbb{R}^{n}.
\end{align}

The control mapping $\mathcal{C}(\cdot)$ transforms state observations into control commands:
\begin{align}
\mathbf{V} = \mathcal{C}(\mathcal{S}) \triangleq \mathcal{C}(\mathbf{P}, \mathbf{P}^{d}, \mathbf{E}, \boldsymbol{\eta}, t),
\end{align}
where $\mathcal{S}=(\mathbf{P}, \mathbf{P}^{d}, \mathbf{E}, \boldsymbol{\eta}, t)$ represents the system state tuple at time $t$.

\subsection{Position-Based Control System}

SDKs of COTS UAV platforms only accept body-frame velocity commands, therefore, to complement our visual-only localization system, we implement a practical position control solution.
Our approach adapts established control principles to bridge this interface constraint.
The controller applies a fundamental coordinate transformation method that converts desired world-frame positions to compatible body-frame velocity commands.
The position and yaw errors for each drone $i$ can be calculated as:
\begin{equation}
\mathbf{e}_i =
\begin{bmatrix}
e_{x,i} \\ e_{y,i} \\ e_{z,i} \\ e_{\psi,i}
\end{bmatrix} =
\begin{bmatrix}
x^d_i - x_i \\ y^d_i - y_i \\ z^d_i - z_i \\ \arg\min_{k \in \{-1,0,1\}} |\psi^d_i - \psi_i + 360k|
\end{bmatrix}.
\end{equation}

To mitigate measurement noise, we implemented first-order filtering on velocity error estimates:
\begin{equation}
\dot{\mathbf{e}}^{f}_{i}(t) = \alpha\dot{\mathbf{e}}^{f}_{i}(t-1) + \beta\dot{\mathbf{e}}_i(t),
\end{equation}
where $\alpha$ and $\beta$ are two hyperparameters.
Then the complete control law is:
\begin{equation}
\mathbf{v}_i =
\begin{bmatrix}
v^b_{x,i} \\
v^b_{y,i} \\
v^b_{z,i} \\
\omega_{\psi,i}
\end{bmatrix} =
\begin{bmatrix}
\mathbf{R}(\psi_i) & \mathbf{0} \\
\mathbf{0} & \mathbf{I}
\end{bmatrix}
\begin{bmatrix}
K_x^p e^{f}_{x,i} + K_x^d \dot{e}^{f}_{x,i} \\
K_y^p e^{f}_{y,i} + K_y^d \dot{e}^{f}_{y,i} \\
K_z^p e^{f}_{z,i} + K_z^d \dot{e}^{f}_{z,i} \\
K_\psi^p e^{f}_{\psi,i} + K_\psi^d \dot{e}^{f}_{\psi,i}
\end{bmatrix},
\end{equation}
where $\mathbf{K}^p = [K_x^p, K_y^p, K_z^p, K_\psi^p]$ represents proportional gains for position errors along the x, y, and z axes and the yaw angle error, while $\mathbf{K}^d = [K_x^d, K_y^d, K_z^d, K_\psi^d]$ represents derivative gains for the corresponding velocity errors.
The yaw angle $\psi_i$ quantifies the angular deviation between the drone's longitudinal axis and the principal reference axis of the SLAM coordinate frame, measured in the horizontal plane. This orientation parameter is directly extractable from our state estimation module of the localization system. $\mathbf{R}(\psi_i)$ represents a rotation matrix that transforms coordinates from world frame to body frame, defined as:
\begin{equation}
\mathbf{R}(\psi_i) =
\begin{bmatrix}
\cos(\psi_i) & \sin(\psi_i) \\
-\sin(\psi_i) & \cos(\psi_i)
\end{bmatrix}.
\end{equation}
It is important to note that this rotation matrix may vary depending on the specific coordinate system definitions adopted in the implementation. The form presented here corresponds to our system configuration, but alternative representations may be required for different coordinate conventions.


Testing demonstrates that this straightforward application of coordinate transformations provides satisfactory position tracking for intended applications, offering a valuable reference for developers working with commercially-constrained UAV platforms where direct position control is unavailable through the provided SDK.

\section{Experiments and Results} \label{sec:experiments}

In this section, we present a comprehensive evaluation of our proposed AirSwarm platform through two key experiments: (1) End-to-End Communication Latency Analysis, (2) Navigation Performance Evaluation. The latency analysis quantifies system responsiveness across control and video pathways, establishing the viability of our architecture for real-time applications. The performance evaluation encompasses the system's navigation capabilities, initialization robustness, localization success rate, and other key metrics.

\subsection{Experimental Platform}

Our experimental platform operates independently using onboard sensors without requiring expensive external positioning equipment such as VICON motion capture systems. The system comprises:
\begin{itemize}
    \item Three DJI Tello EDU drones,
    \item Three Raspberry Pi 4B units (4GB RAM, Ubuntu 20.04) serving as network bridges,
    \item One NVIDIA Jetson AGX Orin (64GB RAM, Ubuntu 22.04) as the central computing unit,
    \item Intel RealSense D455 camera for mapping.
\end{itemize}
Note that the RealSense D455 camera (848×480 resolution, 30fps) is used solely for mapping. During navigation, we only use the onboard monocular camera (960×720 resolution, 30fps) on the drone for localization.




\subsection{Communication Latency Analysis}

\begin{table}[h]
    \centering
    \caption{Control Latency Analysis in (ms).}
        \renewcommand{\arraystretch}{1.2}
    \begin{tabular}{llccc}
        \hline
        \hline
        \textbf{Component} & \textbf{Protocol} & \textbf{Min} & \textbf{Max} & \textbf{Mean} \\ 
        \hline
        PC $\leftrightarrow$ RPi Link & UDP/Ethernet & 0.15 & 1.75 & 0.89 \\ 
        \hline
        RPi Forwarding & IPTABLES  Forward & 0.03 & 0.08 & 0.03 \\ 
        \hline
        RPi $\leftrightarrow$ Tello Link& UDP/Wi-Fi & 4.14 & 66.3 & 25.9 \\ 
        \hline
    \end{tabular}
    \label{tab:control_latency}
\end{table}

We conduct real-time flight tests with DJI Tello drone hovering at a distance of 10 meters from Raspberry Pi and PC for 5 minutes without obstruction. Table \ref{tab:control_latency} presents the latency analysis of control commands transmission between PC, Raspberry Pi (functioning as a forwarding node with iptables), and Tello drone. The results show that latency introduced by wired Ethernet connection (PC$\leftrightarrow$RPi, 0.889ms) and iptables forwarding (0.034ms) is negligible, while the wireless communication between RPi and Tello contributes the majority of command latency (25.886ms). Table \ref{tab:video_analysis} demonstrates the performance metrics of video streaming. The relatively high end-to-end video latency (174.505ms) is primarily attributed to H.264 encoding/decoding process and the bandwidth limitations of the Wi-Fi link, as indicated by the fluctuating bitrate (0.630-4.029 Mbps). Nevertheless, both command and video latencies remain within acceptable bounds for real-time drone control and monitoring applications.

\begin{table}[htbp]
\centering
\caption{End-to-End Video Stream Analysis}
\begin{tabular}{|l|c|c|c|c|}
\hline
\textbf{Metric} & \textbf{Min} & \textbf{Max} & \textbf{Std Dev} & \textbf{Mean} \\
\hline
Latency (ms) & 99.277 & 218.526 & 37.011 & 174.505 \\
\hline
Bitrate (Mbps) & 0.630 & 4.029 & 0.643 & 2.876 \\
\hline
Resolution & \multicolumn{4}{c|}{720p (960×720)} \\
\hline
Frame Rate (fps) & \multicolumn{4}{c|}{30} \\
\hline
Codec & \multicolumn{4}{c|}{H.264} \\
\hline
Transport Protocol & \multicolumn{4}{c|}{UDP} \\
\hline
\end{tabular}
\label{tab:video_analysis}
\end{table}

\begin{table*}[t]
\centering
\renewcommand{\arraystretch}{1.2}
\setlength{\tabcolsep}{5pt}
\caption{Performance Comparison of Multi-Agent SLAM Methods. Our system demonstrates greater robustness, higher success rates, and improved accuracy compared to CCM-SLAM. As CP-SLAM requires RGBD input, which UAVs lack, Depth Anything Model V2 (tiny) was used to estimate depth, leading to higher errors due to model inaccuracies. In contrast, our method operates near real-time, handles rotation and intermittent transmission, and eliminates the need for multi-agent calibration and initialization. }
\label{tab:slam-comparison}
\newcommand{\cmark}{\textcolor{green}{\ding{51}}}
\newcommand{\xmark}{\textcolor{red}{\ding{55}}}
\begin{tabular}{l ccc c c c cc c c}
\toprule
\multirow{1.8}{*}{\textbf{Method}} & \multicolumn{3}{c}{\textbf{APE (cm)}} & \multirow{1.8}{*}{\textbf{FPS}} & \multirow{1.8}{*}{\textbf{Success Rate}} & \multirow{1.8}{*}{\textbf{Communication}} & \multicolumn{2}{c}{\textbf{Support}} & \multirow{1.8}{*}{\textbf{Real-Time}} & \multirow{1.8}{*}{\shortstack{\textbf{Initialization}\\\textbf{Free}}} \\[2pt]
\cmidrule(lr){2-4} \cmidrule(lr){8-9}
 & \textbf{UAV1} & \textbf{UAV2} & \textbf{UAV3} & & & & \textbf{Rotation} & \textbf{Intermittent} & & \\
\midrule
CCM-SLAM\cite{schmuck2019ccm} & Fail & 3.02 & 6.29 & 14.22$\times$3 & 3\% & 0.99 Mbps & \xmark & \xmark & \cmark & \xmark \\[3pt]
CP-SLAM\cite{jiarui2023cpslam} & 61.30 & 63.90 & 67.30 & 1.67 & 31\% & 368.64 Mbps & \cmark & \cmark & \xmark & \xmark \\[3pt]
Proposed & 2.34 & 2.55 & 3.87 & 10.80$\times$3 & 99\% & 1.86 Mbps & \cmark & \cmark & \cmark & \cmark \\
\bottomrule
\end{tabular}
\end{table*}

\subsection{Navigation Performance Evaluation}

\textbf{Task Design:} To demonstrate the capabilities of our system in real-world applications, we designed a coordinated multi-UAV formation task where three Tello drones were commanded to trace the letters ``NTU" in 3D space.
Our experimental protocol involved simultaneous deployment of all three UAVs, each executing predefined trajectories within a common environment. For each method under evaluation, we first generated prebuilt maps using their respective mapping algorithms on identical datasets, then assessed navigation performance during trajectory execution.

\textbf{Baseline Selection:} 
For purely multi-agent visual SLAM systems, the available candidates are limited, as most existing works prioritize Stereo \cite{xu2024d,liu2024omninxt,10321649,zhou2022swarm} or LiDAR-based \cite{zhu2024swarm} solutions for robustness. For visual multi-agent SLAM, we selected CCM-SLAM and CP-SLAM, both of which support shared map usage for collaborative SLAM. While newer variants of CCM-SLAM, such as COVINS \cite{schmuck2021covins,patel2023covinsg}, are available, they lack map reuse capabilities, making direct comparison challenging.
Additionally, we included CP-SLAM \cite{jiarui2023cpslam} in our evaluation, as it represents a collaborative SLAM approach leveraging neural point-based representations. However, CP-SLAM presents two key challenges. First, its open-source implementation was non-functional, requiring us to reimplement the system, which we will release upon paper acceptance. Second, CP-SLAM relies on RGB-D input, which is not standard on commercial off-the-shelf (COTS) drones. To address this, we integrated the Depth Anything Model V2 (tiny) to generate depth maps and conducted offline evaluations for faster tracking performance only. The results are summarized in Tab. \ref{tab:slam-comparison}.

\begin{figure}[htbp]
    \centering
    \subfloat[3D trajectory comparison.]
    {\includegraphics[width=0.22\textwidth]{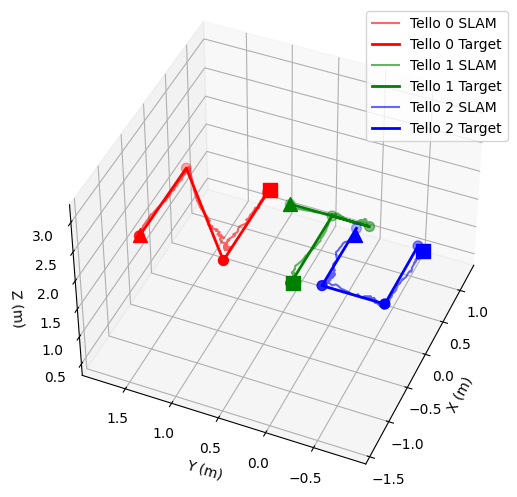}}
    \hfill
    \subfloat[2D trajectory comparison.]
    {\includegraphics[width=0.25\textwidth]{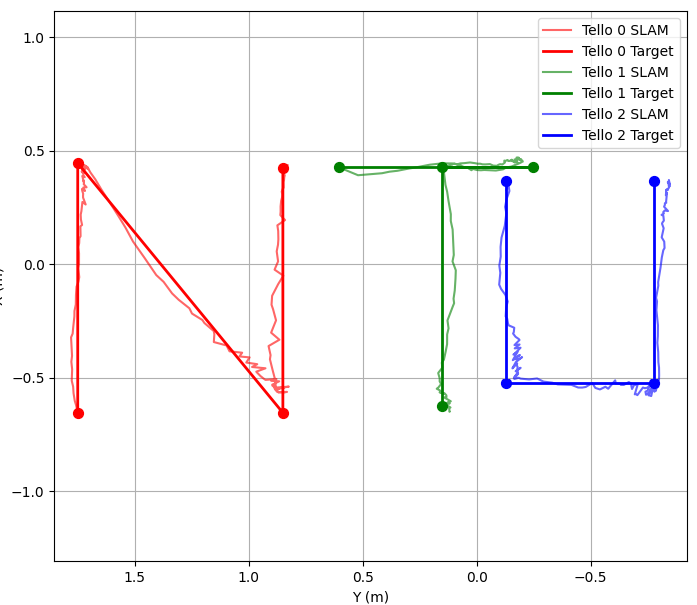}}
    
    \subfloat[Real-time SLAM visualization in Rviz.]
    {\includegraphics[width=0.45\textwidth]{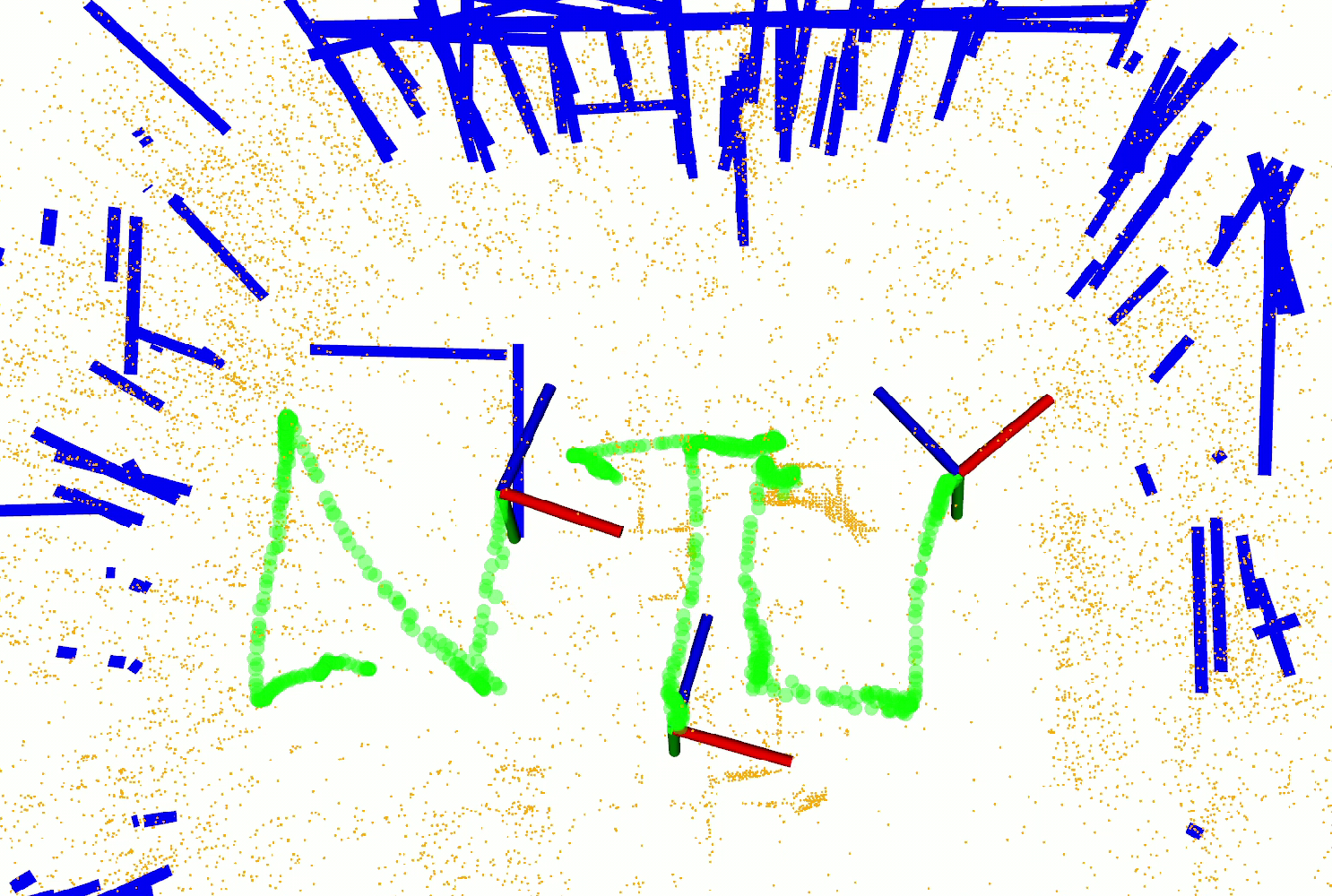}}
    
    \centering
    \subfloat[Experimental environment.]
    {\includegraphics[width=0.45\textwidth]{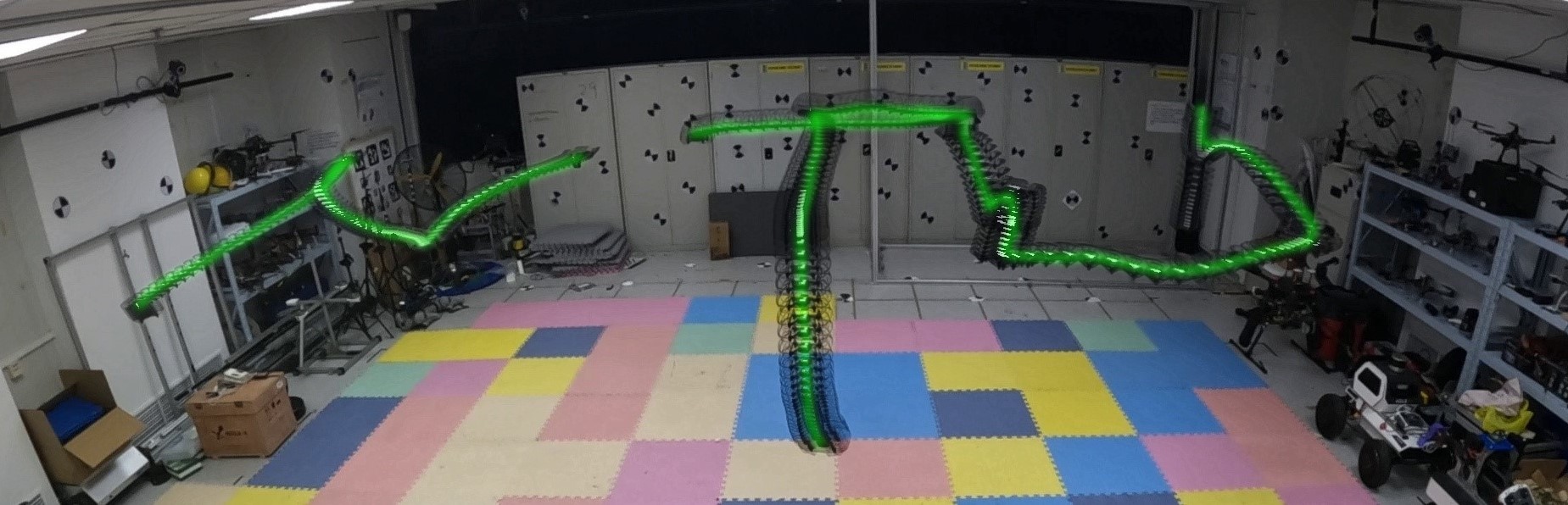}}
    
    \caption{Comparison of SLAM-estimated and Reference Trajectories in Multi-UAV Formation Flight}
    \label{fig:quadrotor}
        \vspace{-15pt}
\end{figure}

\textbf{Results Evaluations:} We present the results in Table \ref{tab:slam-comparison} and Fig~\ref{fig:quadrotor}. Note that the absolute pose errors (APE) \cite{grupp2017evo} in Table \ref{tab:slam-comparison} shows the localization error with ground truth generated by motion capture systems.
The results show that each drone in our system maintained centimeter-level accuracy throughout the flight, with APE of 2.34cm, 2.55cm, and 3.87cm for the drones tracing the letters N, T, and U , respectively. Despite comparable APE metrics in successful trials, alternative approaches exhibited critical operational limitations. CCM-SLAM achieved merely 3\% successful completions across all trials, compared to our system's 99\% success rate. The key issues with CCM SLAM is that the intermittent image transfers cause the system to lose the connection for a short period, which makes it lose connections.



Our system is also highly efficient. During the mapping phase, AirSwarm achieved a processing rate of 30.83Hz using the RealSense D455 camera. In the relocalization phase, the system maintained consistent performance above 10Hz across all three Tello drone streams simultaneously (average 10.80Hz per drone), meeting the real-time requirements for responsive control.
Meanwhile, CP-SLAM's excessive computational demands prevented real-time operation on resource-constrained Jetson platforms, rendering it impractical for edge computing applications despite acceptable accuracy in laboratory settings. Additionally, CP-SLAM's substantial communication overhead (368.64 Mbps) would strain network infrastructure in multi-agent deployments. These comparative results highlight that while competing methods may demonstrate acceptable accuracy in isolated successful cases, they lack the computational efficiency and operational reliability required for consistent real-world deployment.

\begin{figure*}
    \centering
    \includegraphics[width=1\linewidth]{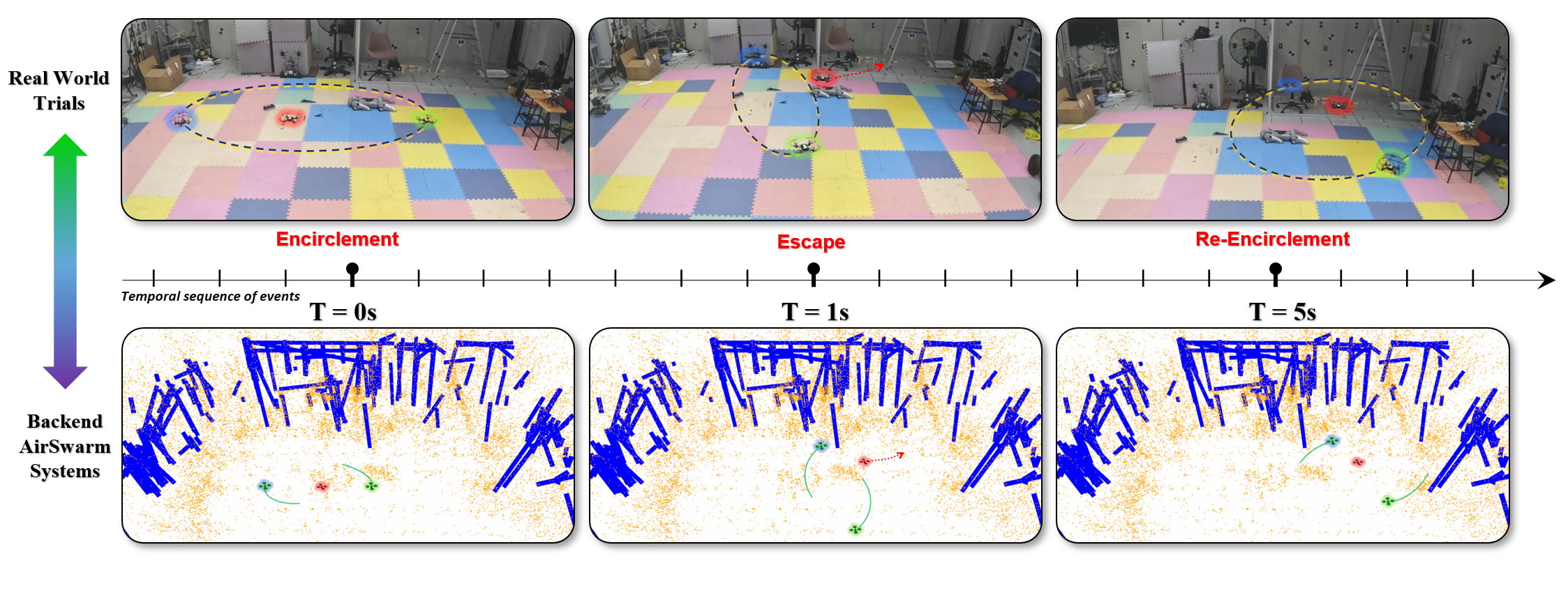}
        \vspace{-25pt}
    \caption{This visualization represents multi-agent aerial tracking and encirlement coordination using the proposed solution.}
    \label{fig:case_study}
        \vspace{-15pt}
\end{figure*}

\begin{figure}
    \centering
    \includegraphics[width=1\linewidth]{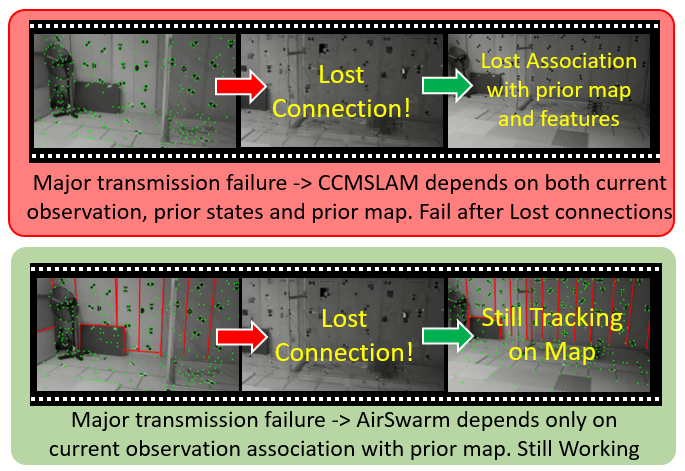}
    \caption{In the presence of communication noise, MAP-based SLAM like CCM-SLAM is more prone to errors due to its dependence on prior states, whereas AirSwarm is based on MLE and demonstrates greater resilience with better noise-handling capabilities. It is the key reason why the proposed solution is better for low-cost COTS swarm research.}
    \label{fig:MLMAP}
\end{figure}

\textbf{Detailed Analysis: } The fundamental algorithmic distinction between these approaches is illustrated in Fig~\ref{fig:MLMAP}, which contrasts the Maximum A Posteriori (MAP) approach used by CCM-SLAM with our Maximum Likelihood Estimation (MLE) approach. The MAP framework incorporates motion model constraints that create interdependencies between sequential pose estimates, requiring precise initialization and continuous tracking to maintain global consistency. CCM-SLAM consequently failed to properly associate the local map with the global reference frame for UAV1 across multiple experimental iterations, confining navigation to local coordinates and resulting in catastrophic trajectory deviation. UAV2 and UAV3 using CCM-SLAM achieved successful localization only after numerous initialization attempts, highlighting the fragility of this tightly-coupled approach.
In contrast, our MLE-based AirSwarm framework establishes direct probabilistic relationships between current camera poses and observations relative to the shared prior map, without enforcing temporal consistency constraints. This architectural decision enables each pose estimate to be derived independently from current observations, conferring inherent resilience against initialization errors and coordinate transformation challenges. As evidenced in our trials, even experience communication failures, the system maintains reliable tracking with respect to the global map.


These characteristics establish our approach as particularly suitable for both educational and research platforms. In educational contexts, the system's moderate processing requirements ensure that algorithmic behaviors remain transparent and interpretable, allowing students to observe fundamental localization concepts in action. For research applications, the framework's resilience to communication interruptions and initialization variability provides a reliable foundation for investigating novel multi-agent coordination strategies, collaborative perception algorithms, and autonomous navigation techniques. The consistent centimeter-level accuracy across varying conditions supports repeatable experimentation, while the computational efficiency enables deployment on resource-constrained platforms typical in both preliminary research investigations and instructional laboratories.

\section{Case Study: Swarm-Based UAV Tracking and Encirclement}
To verify the applicability of the proposed AirSwarm framework in real-world control problems, we evaluate its effectiveness in a multi-UAV autonomous encirclement and re-encirclement task \cite{liu2023non}, as shown in Fig. \ref{fig:case_study}. This application involves a swarm of UAVs dynamically coordinating to track and encircle an adversarial drone using minimal sensing capabilities. The method integrates range-only localization and adaptive anti-synchronization controllers, enabling robust operation in GPS-denied environments. By leveraging AirSwarm’s logically distributed decision-making and multi-agent trajectory planning, we demonstrate that multi-UAV collaboration enhances encirclement efficiency, reducing reaction time in high-speed engagements. 

Traditionally, conducting research in UAV swarm coordination and interception would require custom-built drones equipped with Ultra-Wideband (UWB) modules for precise localization and navigation. These setups are not only expensive but also prone to significant hardware damage when intentional collisions or adversarial interactions occur. The cost of repairing or replacing drones, combined with the complexity of integrating specialized localization hardware, has made such research financially inaccessible for many academic institutions and smaller research labs.

With the proposed AirSwarm framework, we enable a more affordable and cost-effective approach to multi-UAV experimentation. By leveraging commercial off-the-shelf (COTS) drones, lightweight sensing strategies, and logically distributed decision-making, AirSwarm eliminates the need for expensive localization infrastructure while maintaining high experimental fidelity. This affordability makes it an ideal platform for educational and research applications, allowing students and researchers to explore multi-agent aerial coordination, interception strategies, and swarm intelligence without incurring prohibitive costs. Furthermore, the modular nature of AirSwarm ensures scalability, making it adaptable for a wide range of budget-friendly experimental setups, ultimately democratizing access to UAV swarm research.

\section{Limitation and Future Works} \label{sec:limitations}
Despite the promising results demonstrated by AirSwarm, several limitations warrant discussion and point toward future research directions. The primary constraint of the current implementation lies in its computational scalability, as the Nvidia Jetson AGX Orin platform limits simultaneous coordination to three Tello drones. While this limitation could be addressed through more powerful computing hardware, it represents a fundamental trade-off between system cost and swarm size. Additionally, although our localization module demonstrates robust performance across various indoor and outdoor environments under different illumination conditions, its effectiveness diminishes in scenarios with limited point and line features, particularly in textureless environments or areas with repetitive patterns where feature extraction becomes challenging.


Looking forward, these limitations present several promising avenues for future research. The development of more computationally efficient \cite{nguyen2024eigen} visual SLAM algorithms could enable larger swarm formations on existing embedded hardware. Additionally, an intriguing direction involves implementing a multi-center computational architecture where several embedded computing units work cooperatively through task partitioning and load balancing. These targeted advancements would strategically extend AirSwarm's capabilities while preserving its fundamental goal of providing accessible, infrastructure-independent swarm robotics technology that bridges the gap between research prototypes and practical applications.

\section{Conclusion} \label{sec:conclusion}

This paper presents AirSwarm, a novel approach to democratizing drone swarm technology by integrating commercial off-the-shelf (COTS) drones with sophisticated visual SLAM techniques and hierarchical control principles. Our system achieves professional-grade performance with cm-level position tracking accuracy and control latencies under 27ms during complex formation flights, all without relying on expensive external positioning infrastructure.

The significance of this work extends beyond its technical implementation, establishing a new paradigm for accessible multi-robot research and education. By implementing logically distributed processes within a centralized computational framework, the system achieves a 99\% experimental success rate—substantially outperforming comparable approaches that struggled with initialization and communication resilience.

A fundamental contribution of our work is the versatile control framework that operates with virtually any COTS drone equipped with video feedback capabilities. This design enables drift-free visual localization using only onboard cameras, allowing operation across diverse environments without specialized infrastructure.

The architectural contributions provide methodological insights into designing hierarchical perception-action loops for resource-constrained autonomous systems. By balancing performance, accessibility, and usability, AirSwarm establishes a foundation for democratizing access to sophisticated robotics research, potentially accelerating the transition of swarm technologies from laboratory demonstrations to practical field applications across multiple disciplines.


%

\bibliographystyle{./IEEEtran} 
\bibliography{./IEEEabrv,./IEEEexample}

\end{document}